\documentclass[sn-mathphys,Numbered]{sn-jnl}% Math and Physical Sciences Reference Style
\usepackage{graphicx}%
\usepackage{amsmath,mathtools}
\usepackage{multirow}%
\usepackage{amsmath,amssymb,amsfonts}%
\usepackage{amsthm}%
\usepackage{mathrsfs}%
\usepackage[title]{appendix}%
\usepackage{xcolor}%
\usepackage{textcomp}%
\usepackage{manyfoot}%
\usepackage{booktabs}%
\usepackage{algorithm}%
\usepackage{algorithmicx}%
\usepackage{algpseudocode}%
\usepackage{listings}%
\theoremstyle{thmstyleone}%
\theoremstyle{thmstyletwo}%

\usepackage[acronym]{glossaries}

\theoremstyle{thmstylethree}%

\newcommand{\MCTSIts}{\textcolor{black}{Iterations}}

\newcommand{\PercentageNodesF}{Node expansion rate per iteration}

\newcommand{\MostVisitedF}{Most visited node-based result}

\newcommand{\BestRewardF}{Highest reward-based result}

\newacronym{siea}{SIEA-MCTS}{Semantically-Inspired Evolutionary Algorithm Monte Carlo Tree Search}
\newacronym{eamcts}{EA-MCTS}{Evolutionary Algorithm Monte Carlo Tree Search}
\newacronym{mcts}{MCTS}{Monte Carlo Tree Search}
\newtheorem{mydef}{Def.}
\newacronym{amaf}{AMAF}{All-moves-as-first}
\newacronym{ucb1}{UCB1}{Upper Confidence Bounds}

\usepackage{array}
\newcolumntype{P}[1]{>{\centering\arraybackslash}p{#1}}

\begin{document}

 \title[Article Title]{An Analysis on the Effects of Evolving the Monte Carlo Tree Search  Upper Confidence for Trees Selection Policy on Unimodal, Multimodal and Deceptive Landscapes}

%\title[Article Title]{Article Title}

%%=============================================================%%
%% Prefix	-> \pfx{Dr}
%% GivenName	-> \fnm{Joergen W.}
%% Particle	-> \spfx{van der} -> surname prefix
%% FamilyName	-> \sur{Ploeg}
%% Suffix	-> \sfx{IV}
%% NatureName	-> \tanm{Poet Laureate} -> Title after name
%% Degrees	-> \dgr{MSc, PhD}
%% \author*[1,2]{\pfx{Dr} \fnm{Joergen W.} \spfx{van der} \sur{Ploeg} \sfx{IV} \tanm{Poet Laureate} 
%%                 \dgr{MSc, PhD}}\email{iauthor@gmail.com}
%%=============================================================%%

\author*[1]{\fnm{Edgar} \sur{Galv\'{a}n}}\email{edgar.galvan@mu.ie}
\equalcont{Both authors contributed equally to this work.}
\author[2]{\fnm{Fred} \sur{Valdez Ameneyro}}\email{fred.valdezameneyro.2019@mumail.ie}
\equalcont{Both authors contributed equally to this work.}

%\author[1,2]{\fnm{Third} \sur{Author}}\email{iiiauthor@gmail.com}
%\equalcont{These authors contributed equally to this work.}

\affil*[1]{\orgdiv{Naturally Inspired Computation Research Group, Department of Computer Science, Hamilton Institute, National University of Ireland Maynooth, Lero}, \country{Ireland}}

\affil[2]{\orgdiv{Naturally Inspired Computation Research Group, Department of Computer Science, Hamilton Institute, National University of Ireland Maynooth},  \country{Ireland}}

%\affil[3]{\orgdiv{Department}, \orgname{Organization}, \orgaddress{\street{Street}, \city{City}, \postcode{610101}, \state{State}, \country{Country}}}

%%==================================%%
%% sample for unstructured abstract %%
%%==================================%%

\abstract{Monte Carlo Tree Search (MCTS) is a best-first sampling method employed in the search for optimal decisions. The effectiveness of MCTS relies on the construction of its statistical tree, with the selection policy playing a crucial role. A selection policy that works particularly well in MCTS is the Upper Confidence Bounds for Trees, referred to as UCT. The research community has also put forth more sophisticated bounds aimed at enhancing MCTS performance on specific problem domains. Thus, while MCTS UCT generally performs well, there may be variants that outperform it. This has led to various efforts to evolve selection policies for use in MCTS. While all of these previous works are inspiring, none have undertaken an in-depth analysis to shed light on the circumstances in which an evolved alternative to MCTS UCT might prove advantageous. Most of these studies have focused on a single type of problem. In sharp contrast, this work explores the use of five functions of different natures, ranging from unimodal to multimodal and deceptive functions. We illustrate how the evolution of MCTS UCT can yield benefits in multimodal and deceptive scenarios, whereas MCTS UCT is robust in all of the functions used in this work.}


%%\pacs[JEL Classification]{D8, H51}

%%\pacs[MSC Classification]{35A01, 65L10, 65L12, 65L20, 65L70}

\maketitle

\section{Introduction}\label{sec:introduction}

Monte Carlo Tree Search (MCTS) is a powerful sampling method used for uncovering optimal decisions through the execution of random samples within the decision space, ultimately culminating in the construction of a tree based on partial results. The efficacy of MCTS relies directly on the outcomes of its simulations. In theory, with boundless memory and computational resources, the search tree would yield the optimal solution~\cite{kocsis2006bandit}. Nevertheless, in practical, real-world scenarios, MCTS often excels at generating highly satisfactory approximate solutions~\cite{6850585,galvan2014heuristic}.

MCTS has gained popularity in two-player board games, owing much of its success in the complex game of Go~\cite{alphago}, even surpassing professional human players. The application of MCTS extends far beyond gaming, permeating various research areas. For example, it has been extensively investigated in energy-based scenarios~\cite{6850585,galvan2014heuristic} and in the design of deep neural networks' architectures~\cite{wang2019alphax}. These two extreme examples demonstrate the remarkable adaptability, utility, and effectiveness of MCTS across diverse problem domains.

The efficacy of MCTS relies significantly on the construction of its statistical tree. The crucial role in this process is played by the selection policy, particularly when employing the Upper Confidence Bounds for Trees (UCT)~\cite{10.1007/11871842_29}. This requires certain conditions for optimal performance, such as the delicate balance between exploration and exploitation when selecting a child node. Moreover, fine-tuning some of the MCTS UCT's parameters can contribute to an enhanced performance of MCTS.

It is worth mentioning that  more sophisticated bounds in MCTS have been proposed by the research community, including the introduction of a third term to the UCT formula, discussed in Section~\ref{sec:background}, on single-player MCTS scenarios~\cite{DBLP:journals/kbs/SchaddWTU12}. Another variant, UCB1-tuned, modifies the UCT expression to mitigate the impact of the exploration term~\cite{DBLP:journals/ml/AuerCF02}. Thus, it is evident that while MCTS UCT demonstrates proficiency across a range of problems, its adaptation or refinement can have an even more favourable impact on MCTS in diverse problem domains.

As a result of this, multiple scientific studies have been proposed to automatically evolve the MCTS UCT (we discuss relevant studies in Section~\ref{sec:related}). A common denominator in all of these studies is the focus on a single problem, which hinders the ability to shed light on the limitations and potentials of evolving the MCTS UCT in different landscape features with the aim of generalising conclusions. Additionally, a recurring trend in these studies is the use of large population sizes and a substantial number of generations, which limits the practical application of evolved selection policies in MCTS for online simulations.

%In Section~\ref{sec:related}, we discuss the body of research that explores the automatic modification of the MCTS selection policy using Evolutionary Algorithms (EAs)~\cite{EibenBook2003}. Notably, these studies have perfected in on adapting the UCT in diverse contexts, ranging from the game of Ms Pacman~\cite{6633639}, MiniDungeons 2~\cite{DBLP:journals/tciaig/HolmgardGLT19}, to the board game of Carcassonne~\cite{Fred}. While these ingenious approaches, tested on a single game/problem, have provided valuable insights, a common challenge lies in extrapolating their findings to general cases to ascertain when the evolution of a selection policy in MCTS proves advantageous or otherwise.

%Moreover, as mentioned previously, MCTS is incredibly popular thanks to its adaptability and its ability to yield exceptional results, contingent on the execution of sufficient simulations. However, a significant portion of prior research, including these aforementioned studies, often relies on large population sizes and a substantial number of generations. This approach, while insightful, can pose limitations when considering real-time applications.

The main goal of this  work is to shed light on the implications of evolving the UCT in MCTS through the use of EAs, across a spectrum of landscape features. This encompasses a diverse range, spanning Unimodal, Multimodal and Deceptive landscapes, rather than focusing our attention on a particular problem. The main contributions of our work are as follows,
\begin{enumerate}
\item Firstly, instead of confining our focus to a single type of problem, as normally adopted by the community, we use five functions of varying nature and complexity. These range from unimodal, multimodal and encompass also deceptive functions.
\item Secondly, we use a cutting-edge Evolutionary Algorithm (EA), inspired by semantics, to evolve a selection policy to be used in MCTS instead of UCT.
\item Thirdly, departing from the convention of employing large populations and extensive generations, we opt for a modest population size evolved over a limited number of generations. This deliberate choice allows us to explore the potential for successfully evolving a selection policy in lieu of the MCTS UCT. This is facilitated by our semantic-inspired EA method that naturally handles diversity.
\item Fourthly, we conduct a comprehensive comparison involving five variants of the MCTS UCT against our semantic-inspired EA as well as the traditional EAs shedding light on the circumstances under which an evolved selection policy might positively impact MCTS.
\end{enumerate}

The rest of this paper is organised as follows. Section~\ref{sec:related} discusses related work. Section~\ref{sec:background} provides some background in MCTS, EAs, semantics and the functions used in this work. Section~\ref{sec:ai:controllers} discusses in detail the MCTS and EA-based methods used in this work. Section~\ref{sec:experimental} presents the experimental setup. Section~\ref{sec:results} discusses the results obtained by each of the  methods. Finally, Section~\ref{sec:conclusions} draws some conclusions.

\section{Related Work}
\label{sec:related}

{In this section, we discuss the body of research that explores the automatic modification of MCTS using EAs. Notably, these studies have perfected adapting the UCT in diverse contexts, ranging from real-time games like Ms Pacman~\cite{6633639} to board games, such as the Game of Go~\cite{Cazenave2007EvolvingMT}. Next, we discuss some of the most relevant works in evolving parts of the MCTS.
%  While these ingenious approaches tested on a single game or problem have provided valuable insights, a common challenge lies in extrapolating their findings to general cases to ascertain when the evolution of a selection policy in MCTS proves advantageous or otherwise.}

Cazenave~\cite{Cazenave2007EvolvingMT} employed Genetic Programming (GP)~\cite{Koza:1992:GPP:138936}  to evolve the UCT formula for MCTS. His work involved Swiss tournament selection, reproduction, and mutation, with population sizes of 128 or 256 individuals {to evolve formulae that consider a wide variety of alternative statistics and heuristics as terminals, such as the All-moves-as-first (AMAF)  value~\cite{gelly2007mctsrave} of the move, some features of the board, or the best reward among the children of the parent node.} The approach was tested in the game of Go, where it demonstrated superior performance to MCTS and RAVE when a relatively small number of play-outs were used. However, UCT outperformed this approach with a higher number of play-outs.

Inspired by Cazenave's studies, a decade later, Bravi et al.~\cite{Bravi2017EvolvingGU} evolved the MCTS UCB1 formula and tested it using the General Video Game AI framework. They used a population of 100 GP trees, starting with one seeded UCT individual and 99 random trees. Their study considered three scenarios: (a) access to the same information as UCB1, (b) additional game-independent information, and (c) game-specific information. On average, Scenario (c) outperformed the others, including UCB1.

Holmg{\aa}rd et al.~\cite{DBLP:journals/tciaig/HolmgardGLT19} used GP to evolve persona-specific evaluation formulae for MCTS, replacing UCB1 in the deterministic game of MiniDungeons~2. Through selection, crossover, mutation, elitism, and an island model, the authors evolved 100 GP individuals over 100 generations. The evolved personas demonstrated more efficient gameplay compared to UCB1 agents.

{Besides evolving the selection step of MCTS, EAs have been also employed to improve MCTS's simulation step. An example of such contributions} comes from Alhejali and Lucas~\cite{6633639}, who used a GP system to enhance MCTS with heavy rollouts in the game of Ms. Pac-Man. Using multiple functions, including comparison, conditional, and logical operators, they evolved 100 or 500 individuals over 50 or 100 generations. The latter case took approximately 18 days to complete. Later, Lucas et al.~\cite{Lucas2014} used an EA to control parameters that guide the rollouts in MCTS. Their approach demonstrated significant improvements over vanilla MCTS using benchmark problems like Mountain Car and a simplified version of Space Invaders.

{Benbassat and Sipper proposed the EvoMCTS algorithm~\cite{benbassat2013evomcts,benbassat2014evomcts}. The authors proposed the evolution of board evaluation functions using GP. In this approach, the evolved board evaluation function serves to guide the default policy. During the simulation step of MCTS, the default policy selects the best available action by considering a one-step look-ahead evaluation of the next available states. To handle large branching factors and to speed the simulations up, the default policy in EvoMCTS randomly samples a predefined number of actions from each state.}

Baier and Cowling~\cite{8490403} proposed a method to address prohibitive branching factors in MCTS using EAs in Hero Academy{, a card game with complex turns. In their method, each single turn is modelled as a sequence of micro-actions that are methodically mutated using MCTS. Each MCTS node represents a full turn, and each edge represents a potential mutation to that sequence. They incorporate the Bridge Burning technique~\cite{justesen2017oep,schadd2012spmcts} that divides the search into $n$ phases specified by the user. Once the budget for each phase is exhausted, the best child of the root becomes the new root of the statistical tree. All nodes that are not part of the new statistical tree are discarded, allowing the search to proceed.}

A significant portion of prior research, including these aforementioned studies, often relies on early exposure to the application domain to evolve aspects of the MCTS algorithm by using large population sizes and a substantial number of generations. Then, the evolved solution is summoned on the domain and results are collected. This approach, while insightful, can pose generality limitations. For instance, within the same domain, the necessities of the game may greatly vary. As an example, a player in Ms-Pacman requires a different set of skills when the power-ups are collected or nearing the end of a level. %In this work, we opt for a modest population size evolved over a limited number of generations initialised from scratch on each decision to be made by the agent. This deliberate choice allows us to explore the potential for successfully evolving a selection policy in lieu of the MCTS UCT online.

In our previous work~\cite{9659930}, we presented initial results on evolving the UCT using EAs in the game of Carcassonne. We employed two variants of EAs, referred to as EA partially integrated in MCTS and EA in MCTS. This work served as the foundation for our most recent IEEE Trans. on Games article~\cite{9872022}, which employed a state-of-the-art EA inspired by semantics to evolve MCTS UCT in the game of Carcassonne, using a small population and the use of few generations. We demonstrated how this approach yielded results that were either superior or competitive compared to well-known tuned MCTS UCT variants, as well as other MCTS variants tailored for this specific game.

\section{Background}
\label{sec:background}

\subsection{The Mechanics Behind MCTS}

The MCTS algorithm relies on two fundamental principles: (a) the ability to approximate the true value of an action through simulations, and (b) the use of these values to adjust the policy towards a best-first strategy. This iterative process involves constructing a partial tree, guided by the results of prior exploration. The algorithm continues building the tree until a predefined condition is met, such as a specified number of iterations or a designated time for Monte Carlo simulations. The search is then halted, and the best-performing action is executed. In the tree structure, each node represents a state, with directed links leading to child nodes representing actions leading to subsequent states. As with many AI techniques, MCTS exhibits various variants. The steps commonly accepted in MCTS, as outlined in~\cite{6145622}, include:
  
\begin{enumerate}
\item \textit{Selection}: A recursive selection policy guides the descent through the tree until an expandable node is reached. A node is classified as expandable if it represents a non-terminal state and has unvisited child nodes.
\item \textit{Expansion}: Typically, one child is added to expand the tree based on available actions.
\item \textit{Simulation}: From the newly added nodes, a simulation (also called rollout or playout) is conducted to obtain an outcome, such as a reward value.
\item \textit{Back-propagation}: The outcome obtained from the simulation step is propagated backwards through the selected nodes to update their corresponding statistics.
\end{enumerate}

Iterations in MCTS start from the root state (e.g., the current position) and encompass two stages. Initially, when the state is integrated into the tree, a tree policy is employed to select actions. The selection step is a key element and will be discussed in greater detail later in this section. A default policy is employed to carry out simulations to completion, otherwise. This is depicted in Fig.~\ref{fig:mcts}.

One key advancement in MCTS efficiency was the selection mechanism introduced in~\cite{10.1007/11871842_29}. The core concept behind this mechanism was to devise a Monte Carlo search algorithm with a low probability of error if prematurely halted, and one that eventually converges to the optimal solution given sufficient time. In essence, it achieves a balance between exploration and exploitation, a concept elaborated in the following paragraphs.

\begin{figure}[tb]
  \centering
\includegraphics[width=00.75\columnwidth]{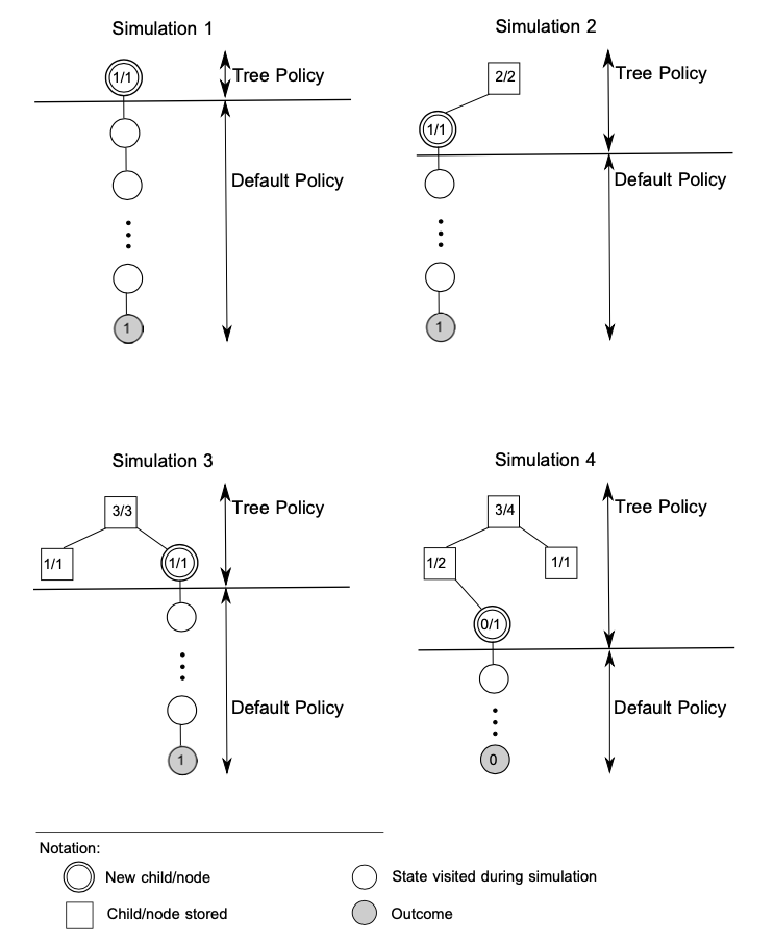}
\caption{Four MCTS iterations, also often referred to as simulations. In Simulation 1, a new node is added to the tree search and its statistics are updated (e.g., outcome and number of visits). Simulation 2, stores the first node and adds a new one. Simulation 3 adds a new node according to the other action (in this case there are only two actions, as indicated by the number of children). Simulation 4 selects (in this case it is a tie, so a random selection is used) and beneath the selected node, a new one is added. Redrawn from~\cite{GELLY20111856}.}
\label{fig:mcts}
\end{figure}

\subsection{Upper Confidence Bounds for Trees}

As mentioned earlier, MCTS operates by approximating the `real' values of possible actions from the current state, accomplished through the construction of a decision tree. The effectiveness of MCTS relies heavily on the strategy used to build this tree, with the selection process playing a key role. One particularly reliable selection mechanism is the UCB1 tree policy~\cite{10.1007/11871842_29}, which is formally defined as:

\begin{equation}
  UCT = \overline{X}_j + C \sqrt{\frac{2 \cdot ln \cdot n}{n_j}}
  \label{eq:uct}
  \end{equation}

\noindent where $n$ is the number of times the parent node has been visited, $n_j$ is the number of times child $j$ has been visited and $C > 0$ is a constant. In case of a tie for selecting a child node, a random selection is normally used~\cite{10.1007/11871842_29}.

This selection mechanism thrives on its ability to strike a balance between exploitation (as seen in the first part of Eq.~\ref{eq:uct}) and exploration (as seen in the second part of Eq.~\ref{eq:uct}). Whenever a node is visited, the denominator of the exploration component increases, resulting in a decrease in its overall contribution. Conversely, if another child node of the same parent is visited, the numerator increases, thereby augmenting the exploration values of unvisited children. The exploration term in Eq.~\ref{eq:uct} ensures that each child node maintains a selection probability greater than zero, a crucial aspect given the stochastic nature of the play-outs.

\subsection{Evolutionary Algorithms}

Evolutionary Algorithms (EAs)~\cite{EibenBook2003} encompass a class of stochastic optimisation algorithms inspired by biological evolution. They employ evolutionary principles to construct adaptive systems that excel at solving a wide range of problems. Notably, their flexibility allows practitioners to blend elements from various EA techniques, erasing the distinct boundaries between these approaches. This facilitates the emergence of a more holistic EA framework, as exemplified in this research.

EAs operate on a population of $\mu$-encoded potential solutions to a specific problem. Each potential solution, referred to as an individual, represents a point within the search space where the optimal solution is situated. Genetic operators are then applied to evolve this population over several generations, aiming to produce improved results for the given problem. Each individual is assessed using a fitness function to ascertain its suitability for the problem. The fitness value assigned to an individual probabilistically influences its likelihood of passing on (part of) its code to subsequent generations.

The evolutionary process is guided by genetic operators, with selection, crossover, and mutation being the core ones employed in most EAs. The selection operator is responsible for picking one or more individuals from the population based on their fitness values. Various selection operators have been proposed. One of the most popular selection operators is tournament selection where the best individual is selected from a pool, normally of size = $[2-7]$, from the population. The stochastic crossover operator, also known as recombination, swaps genetic material between two selected individuals, playing a key role in exploiting the search space. The stochastic mutation operator introduces random changes to an individual's genes, ensuring diversity in the population and recovering potentially lost genetic material during evolution. This evolutionary process continues until a stopping condition is met, such as reaching a maximum number of generations. At this stage, the population comprises the best-evolved potential solutions to the problem and may also represent the global optimal solution.

The field of EAs traces its roots back to four seminal evolutionary methods: Genetic Algorithms~\cite{DBLP:books/aw/Goldberg89}, Evolution Strategies~\cite{Rechenberg10.1007/978-3-642-83814-9_6}, Evolutionary Programming~\cite{Fogel1966}, and Genetic Programming~\cite{Koza:1992:GPP:138936}. In this work, we provide a brief overview of the two methods employed: Evolution Strategies and Genetic Programming. For a more comprehensive understanding, the first author's work in Neuroevolution in Deep Neural Networks~\cite{9383028} offers an insightful summary of all these EAs.

\subsubsection{Evolutionary Algorithm: Genetic Programming (GP)}
GP popularised by Koza~\cite{Koza:1992:GPP:138936}, is a variant of automated programming. In GP, individuals are randomly generated using functional and terminal sets tailored to address a specific problem. The literature presents various types of GP, but the prevalent form within EAs is characterised by a tree-like structure.

\subsubsection{Evolutionary Algorithm: Evolution Strategies (ES)} 

Introduced by Rechenberg in the 1960s~\cite{Rechenberg10.1007/978-3-642-83814-9_6}, ES find application in optimising real-valued representations of problems. In ES, mutation serves as the primary operator, while crossover is an optional, secondary one. Historically, two fundamental forms of ES emerged: the ($\mu$, $\lambda$)-ES and the ($\mu$ + $\lambda$)-ES. Here, $\mu$ denotes the size of the parent population, while $\lambda$ represents the number of offspring generated in the subsequent generation before undergoing selection. In the former ES, offspring replace parents, while in the latter variant, selection is applied to both offspring and parents to form the subsequent generation's population.

\subsection{Semantics}

For clarity purposes, we first briefly give some definitions of semantics, based on the first author's work~\cite{DBLP:conf/gecco/GalvanS19}, that will allow us to describe our approach later in Section~\ref{sec:ai:controllers}.

Let $p \in P$ be a program from a language $P$. When $p$ is applied to an input  $in \in I$, $p$ produces an output $p(in)$. 

\begin{mydef}
\indent  The semantic mapping function $s: P \rightarrow S$  maps any program $g$ to its semantics $s(g)$.
\label{def:general}
\end{mydef}

This means, $\textcolor{black}{s(g_1) = s(g_2) \Longleftrightarrow \forall in >\in I : g_1(in) = g_2(in).}$ The semantics specified in Def.~\ref{def:general} has three properties. Firstly, every program has only and only one semantics. Secondly, two or more programs can have the same semantics. Thirdly, programs that produce different outputs have different semantics. Def.~\ref{def:general} is general as it does not specify how semantics is represented. This work is inspired by a popular version of semantics GP where the semantics of a program is defined as the vector of output values computed by this program for an input set (also known as fitness cases). The latter are not available in MCTS. We then extrapolate this idea to the fitness space. Thus, assuming we use a finite set of iterations, as normally adopted in MCTS, we can now define, without losing the generality, the semantics of a program in the simulations.

\begin{mydef}
  The semantics $s(p)$ of a program $p$ is the vector of values from each independent simulation $sim$.
\label{def:semantics}
  \end{mydef}

Thus, we have that the semantics of a program in MCTS is given by $s(p) = [p(sim_1), p(sim_2), \cdots, p(sim_l)]$,  where $l = |I|$ is the number of fitness iterations.

Based on Def.~\ref{def:semantics}, we can define the \textit{Sampling Semantics Distance} (SSD) between two programs $(p,q)$. That is, let $P$ = $\{p_1, p_2, ..., p_N \}$ and \sloppy{$Q$ = $\{q_1, q_2, \cdots, q_N \}$} be the sampling semantic of Program 1 ($p_1$) and Program 2 ($p_2$) on the same set of sample points, then the SSD between $p_1$ and $p_2$ is defined as

\begin{equation}
  SSD(p,q) = (|p_1-q_1|+|p_2-q_2|+...+|p_N-q_N|)/N
\label{eq:ssd}
\end{equation}

\noindent where $S_p = \{p_1,...,p_N\}$ and $S_q = \{q_1,...,q_N\}$ are the SS of programs $p$ and $q$ based on simulations.

We are now in a position to use the well-known semantic similarity (SSi) proposed by the first author and colleagues~\cite{DBLP:journals/gpem/UyHOML11}. This indicates whether the SSD, shown in Eq.~\ref{eq:ssd}, between two programs, lies between a lower bound $\alpha$ and an upper bound $\beta$ or not. It determines if two programs are similar without being semantically identical. The SSi of two programs $p$ and $q$ on a domain is formally defined as

\begin{equation}
  \textcolor{black}{  SSi(p,q) = (\alpha < SSD(p,q) < \beta)}
  \label{eq:ssi}
\end{equation}

\noindent where $\alpha$ and $\beta$ are the lower and upper bounds for semantic sensitivity, respectively. In our work, we set these as 5 and 10, respectively.

\section{AI Controllers}
\label{sec:ai:controllers}

\subsection{Monte Carlo Tree Search}

The fundamental concept behind MCTS revolves around its four phases: selection, expansion, rollout, and backpropagation. These stages collectively constitute a single {iteration}. Upon the culmination of all iterations, the node with the highest action value is typically selected, aligning with the approach adopted in this study. For a comprehensive understanding of the MCTS algorithm, please refer to the detailed explanation provided in Section~\ref{sec:background}.

\subsection{Evolutionary Algorithm in MCTS}

This EAs-based controller  aims to evolve the mathematical expression employed during the selection phase of MCTS. {To enhance the adaptability of MCTS, we recently introduced the Evolutionary Algorithm Monte Carlo Tree Search (EA-MCTS)~\cite{9659930}. Other works attempt to evolve the tree policy offline to then use the evolved formula in subsequent decisions~\cite{Bravi2017EvolvingGU,benbassat2013evomcts}. EA-MCTS differs from them as it evolves the UCB1 formula online from the ground for every single decision, building the foundations for a more generic EA-based MCTS variant.}

EA-MCTS modifies the selection step of the MCTS algorithm, which is responsible for traversing the statistical tree. When all the children of the root node have been expanded, the EA in the EA-MCTS algorithm kicks in. Our EA employs a ($\mu$,$\lambda$)-ES style of population handling (please, refer to Section~\ref{sec:background}). Once the evolutionary process is finished, the formula returned by the evolutionary process replaces UCT as the tree policy for all subsequent MCTS iterations, which operate as usual. 

The steps of our approach are outlined in Alg.~\ref{alg:mcts_ea}. Initially, we employ the UCT formula as a parent (Line 4) to generate offspring later on (Line 7). In each MCTS call, a newly evolved solution is constructed from scratch. Similar to~\cite{james2017analysis}, ``in the simulation phase, actions are executed uniformly randomly until a terminal state is encountered, at which point some reward is received. Let $f$ be the function and $c$ be the midpoint of the state reached by the rollout. At iteration $t$, a binary reward $r_t$ is drawn from a Bernoulli distribution $r_t \sim Bern (f (c))$''. These rewards are then used to update the MCTS statistical tree (Line 11), starting from the selected node up to the root, including all nodes along the branch. {We run 30 dedicated MCTS iterations (Line 9), referred to as {fitness iterations}, to assign their average reward as the fitness of the evolved expression (Line 14).}

\begin{algorithm}[tbh!]
  \small
\caption{\textcolor{black}{EA-MCTS}}
  \begin{algorithmic}[1]
        \State \noindent \textbf{Input:} Number of gen. $G$, lambda $\lambda$, {fitness iterations} $S$, statistical tree $T$\\
      \noindent \textbf{Output:} Evolved tree policy 
    \Procedure{Evolving\_UCT\_EA}{}
          \State P $\gets$ UCT formula
          \For {$g \gets 0,  \cdots, G$}
               \For {$i \gets 0, \cdots, \lambda$}
               \State O$_i$ $\gets$ subtree\_mutation(P)
               \State a\_fitness $\gets$ 0 
                   \For {$s \gets 0, \cdots, S$}
                      \State temp\_fit(O$_i$) $\gets$ select $T(S)$ and rollout(O$_i$) 
                      \State update $T$
                      \State a\_fitness(O$_i$) $\gets$ a\_fitness(O$_i$) + temp\_fit(O$_i$)
                   \EndFor
                   \State fitness(O$_i$) $\gets$ a\_fitness(O$_i$) / S
              
                   \EndFor
           \EndFor
           \State return P
           \EndProcedure
    \end{algorithmic}
\label{alg:mcts_ea}
\end{algorithm}

\subsection{Semantically-inspired Evolutionary Algorithm in MCTS}
\label{subsection:EA-p-MCTS}

A potential limitation of using a small population size, a contribution of this paper as outlined in Section~\ref{sec:introduction}, is the risk of limited diversity leading to suboptimal performance. To mitigate this, we draw inspiration from semantics to encourage diversity (see Section~\ref{sec:background}). This approach, introduced as Semantically-Inspired Evolutionary Algorithm Monte Carlo Tree Search (SIEA-MCTS)  in our recent work~\cite{9872022}, aims to increase diversity leading to potentially increase performance.

Semantics is a well-known concept in the field of GP~\cite{DBLP:journals/gpem/UyHOML11}. The main idea behind SIEA-MCTS is to use semantics to select the best offspring to serve as a parent in the next generation. This is achieved by computing the semantic distance~\cite{DBLP:conf/gecco/GalvanS19} between the offspring and the parent. This approach is based on the assumption that a semantically different individual, yet sufficiently close to its parent, tends to yield superior results~\cite{DBLP:journals/gpem/UyHOML11}.  Moreover, this area is incredibly active in multiple area including in multi-objective optimisation~\cite{Galvan-Lopez:2016:PPSN,DBLP:conf/micai/LopezVT16,9308386,GALVAN2022108143,9504860}. 

\begin{algorithm}[tbh!]
    \small
	\caption{\textcolor{black}{SIEA-MCTS}}
  	\begin{algorithmic}[1]
          \State \noindent \textbf{Input:} Number of gen. $G$, lambda $\lambda$, {fitness iterations} $S$, statistical tree $T$\\
        \noindent \textbf{Output:} Evolved tree policy 
	    \Procedure{Evolving\_UCT\_EA\_SIEA}{}
            \State P $\gets$ UCT formula
            \For {$g \gets 0,  \cdots, G$}
                 \For {$i \gets 0, \cdots, \lambda$}
                 \State O$_i$ $\gets$ subtree\_mutation(P)
                 \State a\_fitness $\gets$ 0 
                     \For {$s \gets 0, \cdots, S$}
                        \State temp\_fit(O$_i$) $\gets$ select $T(S)$ and rollout(O$_i$) 
                        \State update $T$
                        \State a\_fitness(O$_i$) $\gets$ a\_fitness(O$_i$) + temp\_fit(O$_i$)
                     \EndFor
                     \State fitness(O$_i$) $\gets$ a\_fitness(O$_i$) / S
                
                     \EndFor

                        \State  Sem\_Sel(O,P) \textbf{procedure}

             \EndFor
             \State return P
             \EndProcedure
         
    	\\\hrulefill
\State \noindent \textbf{Input:} Population offspring $O$, Parent $P$\\
        \noindent \textbf{Output:} Best program based on fitness and semantics
             \Procedure{Sem\_Sel}{O,P}
             \State H$_f$ $\gets$ max(fitness($O$))
            \If {More than one offspring from $O$ equals H$_f$}
                %\If{Semantics is in Decision-Policy}
                    \State SSD $\gets $Sampling\_sem\_dist($O,P$)
                    \State SSi $\gets$ Sel. ind(s) within sem. sim. range ($\alpha$, $\beta$)
                    \If{more than one individual within range}
                        \State New\_P $\gets$ individual closest to lower-bound $\alpha$
                    \Else
                        \State New\_P $\gets$ random($O$)
                    \EndIf

                    \Else
                    \State New\_P $\gets$ random($O$)
            \EndIf
            \State \textbf{return} New\_P
             \EndProcedure
    	\end{algorithmic}
	\label{alg:mcts_siea}
\end{algorithm}

{The steps of our approach are outlined in Alg.~\ref{alg:mcts_siea}. We select the offspring based on semantics to serve as a parent in Line 16 where SIEA-MCTS and EA-MCTS differ. The semantically-inspired selector (Line 23) initially retrieves the offspring with the highest fitness, denoted as H$_f$ (Line 24).} If there is more than one offspring with H$_f$ (Line 25), we compute the sampling semantic distance from each offspring in relation to the Parent (Line 27). Subsequently, we calculate the semantic similarity metric using predefined thresholds (Line 27). If multiple individuals from the offspring population fall within this threshold, defined by $\alpha$ and $\beta$, then we select the individual closest to the  $\alpha$ value (Line 29). Otherwise, we randomly select an individual from the offspring population (Line 31).

Our proposed method aims to evolve mathematical expressions that can outperform or at least be on par with UCT in producing competitive results. Thus, ES are employed during the selection step in MCTS. Once a node is chosen by our evolved expression, we proceed to compute the fitness of the evolved expression through rollouts, following the methodology employed in MCTS.

\section{Experimental Setup}
\label{sec:experimental}

\subsection{Test Functions}

The use of benchmark problems has been instrumental in enabling the research community to rigorously assess, validate, and elucidate a wide array of AI methodologies. In this study, we embrace the use of five distinct test functions~\cite{DBLP:conf/ssci/AmeneyroG22}, each characterised by varying degrees of complexity and representing diverse search landscapes. As indicated in Section 1, this constitutes another noteworthy contribution from our work. Rather than exclusively focusing on a single problem for the exploration and elucidation of the evolution of selection policies in MCTS, we leverage these functions to ensure that our evaluations are not contingent on a specific problem domain. 

Our selection comprises five functions, each exhibiting a different level of complexity, spanning from unimodal functions to multimodal functions, and culminating in highly deceptive functions. We confine both the domain and range of these functions to the interval $[0,1]$. These functions are visually shown in Fig.~\ref{fig:functions}.

\begin{figure}[t] 
    \centering\includegraphics[width=0.75\columnwidth]{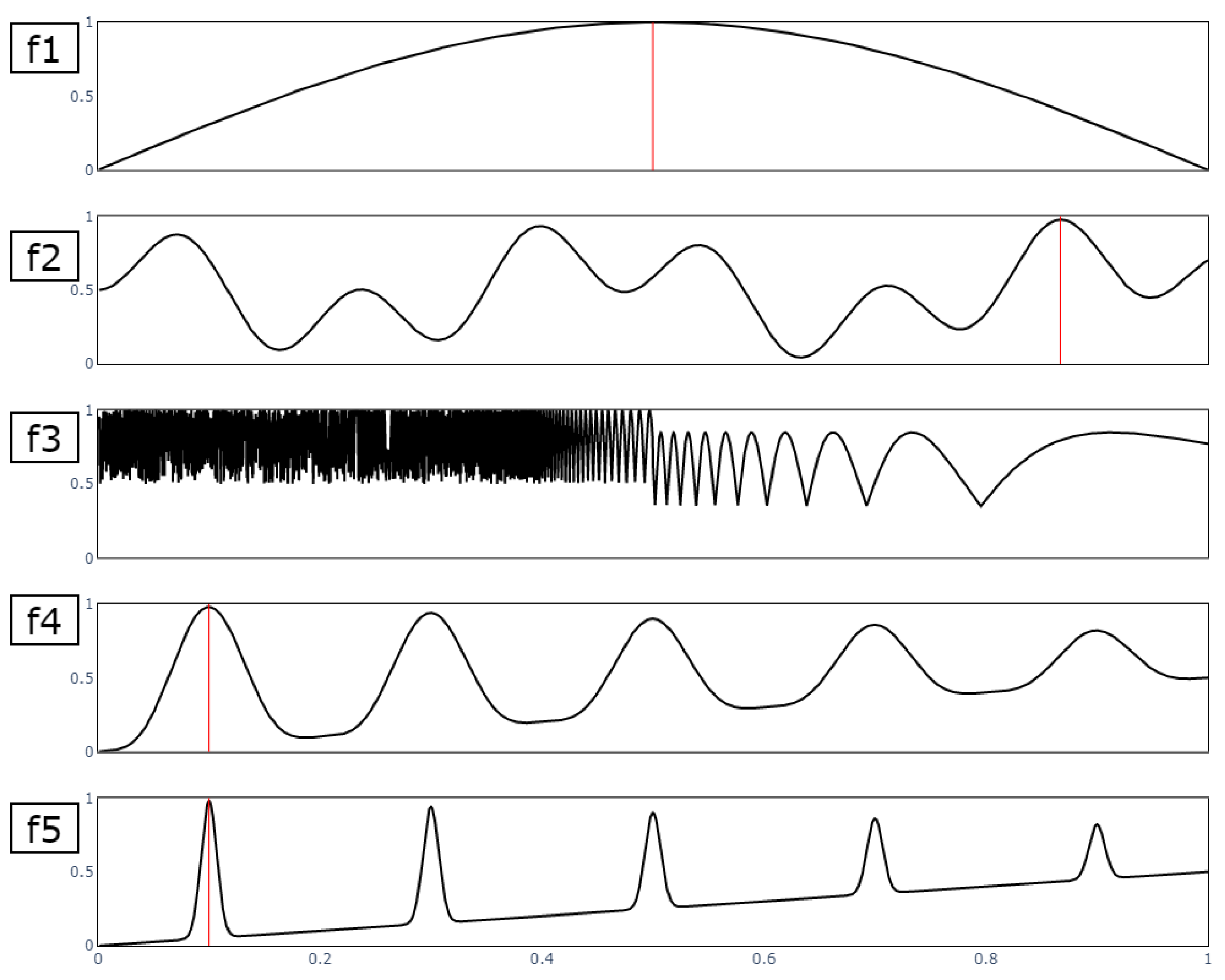}
    \caption{Five functions of different degrees of difficulty used in this study, including an unimodal function (f$_1$), multimodal functions (f$_2$ and f$_3$) and deceptive functions (f$_4$ and f$_5$).}
    \label{fig:functions}
\end{figure}

The first function, shown in Fig.~\ref{fig:functions}, f$_1$, depicts an unimodal function. The global optima is shown with a red vertical line. The function is defined by Eq.~\ref{eq:f1}, 

    \begin{equation}
      f_{1}(x) = \sin(\pi x)
      \label{eq:f1}
    \end{equation}

    The second function, shown in Fig.~\ref{fig:functions}, f$_2$, is a multimodal function, with a global optima situated on the right-hand side of the figure as denoted by the red vertical  line. The function is defined by Eq.~\ref{eq:f2},

    \begin{equation}
      f_{2}(x) =  0.5 \sin(13 x)  \sin(27 x) + 0.5
      \label{eq:f2}
    \end{equation}

   The third function, inspired by~\cite{james2017analysis}, is shown in Fig.~\ref{fig:functions}, f$_3$. It shows a function with a high degree of ruggedness (left-hand side) containing multiple global optimum (not indicated as done with f$_1$ and f$_2$ given the nature of f$_3$). In sharp contrast, the function also shows  `smoothness' on the right-hand side. The function is defined by Eq.~\ref{eq:f3},

    \begin{equation}
        f_{3}(x)=\begin{cases*}
            0.5 + 0.5 | \sin( \frac{1}{x^{5}})| & when $x<0.5$,\\
            \frac{7}{20} + 0.5 | \sin( \frac{1}{x^{5}})| & when $x\geq0.5$.
        \end{cases*} 
        \label{eq:f3}
    \end{equation}

    The fourth function used in this work is shown in Fig.~\ref{fig:functions}, f$_4$. This is a deceptive function, where the global optimum is depicted by the red vertical line in the left-hand side of the figure. This function is defined by Eq.~\ref{eq:f4},
      
    \begin{equation}
        f_{4}(x) =  0.5 x + (-0.7 x+1)\sin(5 \pi  x)^{4}
        \label{eq:f4}
    \end{equation}

    The fifth and final function used in this work is shown in Fig.~\ref{fig:functions}, f$_5$. This is also a deceptive function,  harder compared to f$_4$. This function is defined by Eq.~\ref{eq:f5},
   
    \begin{equation}
        f_{5}(x) =  0.5 x + (-0.7 x+1)\sin(5 \pi  x)^{80}
        \label{eq:f5}
    \end{equation}

These functions, commonly known as function optimisation as described in \cite{bubeck2010x,james2017analysis}, is a problem where each state $s$ represents a domain [$a_{s}$,$b_{s}$], starting at [0,1]. The available actions from state $s$ are $b$ evenly spaced partitions of the domain, sized $\frac{a_{s}-b_{s}}{b}$ each, where $b$ is the selected branching factor. The objective is to find the state where the global maxima of a given function $f$ lies. A state is said to be terminal if its domain is smaller than the threshold $t$, in other words, $b_{s}-a_{s}<t$. The threshold is set as $t=10^{-5}$ and the branching factor is set as $b=2$.

The MCTS rollouts use a random uniform default policy. When a terminal state is reached, $f$ is evaluated at the state's central point $c_{s}=\frac{(a_{s}-b_{s})}{2}$. The reward $r_{s}$ can either be 1 or 0 and is sampled from a Bernoulli distribution $r_{s}\sim Bern(f(c_{s}))$. Thus, $0\leq f(x)\leq 1 | x \in [0,1] $ is ensured for every function $f$.

\subsection{Function and Terminal Sets}

As indicated in Section~\ref{sec:ai:controllers}, we use GP to evolve mathematical expressions that can be used in lieu of the UCT1 (see Section~\ref{sec:background}). To this end, we define the terminal and function sets as follows. The terminal set is defined by  $T =  \{Q(s,a),N(s),N(s,a),C\}$, where $N(s)$ is the number of visits to the node from the MCTS search tree, $N(s,a)$ is the number of visits to a child node, $Q(s,a)$ is the child's node action-value and $C$ is the exploration-exploitation constant. When $C$ is chosen to be mutated, it can take a random value from the following set $r = \{0.5, 1, \sqrt2, 2, 3\}$. The function set is defined by $F = \{+,-,*,\div,\log,\sqrt { }\}$, where the division operator is protected against division by zero and will return 1 for any divisor less than 0.001. Similarly, the natural log and square root operators are protected by taking the absolute values of input values. The values used for all the controllers used in this work are shown in Table~\ref{tab:parameters}.

\begin{table}[h]
\caption{MCTS agents' parameters.}
\begin{tabular}{@{}lr@{}}
\toprule
Parameter & Value  \\
\midrule
\multicolumn{2}{c}{All MCTS agents} \\ \hline
C & {$\frac{1}{2}, 1, \sqrt2, 2, 3$} \\ \hline
Rollout playouts & 1 \\ \hline
\MCTSIts & {5000} \\ \hline

\multicolumn{2}{c}{EA methods}\\ \hline
($\mu$,$\lambda$)-ES &  $\mu=1$, $\lambda=4$  \\ \hline
Generations & 20 \\ \hline
Type of Mutation & Subtree (90\% internal node, 10\% leaf) \\ \hline
Initialisation Method & Initial formula: UCB1 \\ \hline
Maximum depth & 8 \\ \hline
\multirow{2}{*}{Iterations} & 2,430 fitness iterations + 2,570 iterations \\ %\cline{2-2}
  & \textcolor{black}{2,430 fitness iterations + 5,000 iterations}\\ 

\botrule
\end{tabular}
%\footnotetext{Source: This is an example of table footnote. This is an example of table footnote.}
%\footnotetext[1]{Example for a first table footnote. This is an example of table footnote.}
%\footnotetext[2]{Example for a second table footnote. This is an example of table footnote.}
\label{tab:parameters}
\end{table}

\section{Discussion of Results}
\label{sec:results}

\subsection{Exploration vs. Exploitation}

\begin{figure}[tb] 
  \centering\includegraphics[width=0.9\columnwidth]{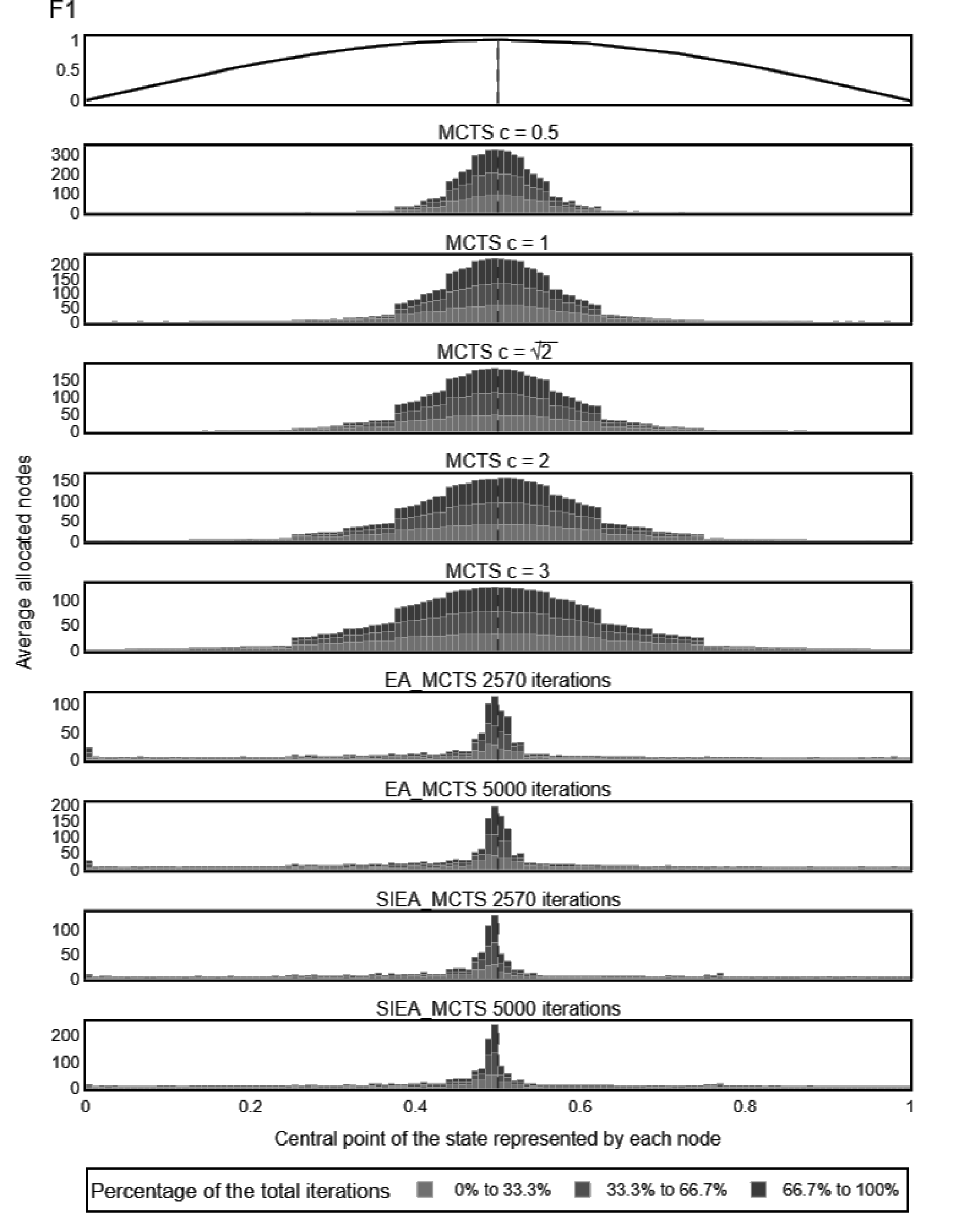}%{Images/f0.eps}
   \caption{Histogram of the location of the nodes using MCTS UCT C = \{0.5,1,$\sqrt{2}$,2,3\} and SIEA-MCTS for f$_1$ (see Section~\ref{sec:background}).}
    \label{fig:f1}
\end{figure}

\begin{figure}[tb] 
  \centering\includegraphics[width=0.9\columnwidth]{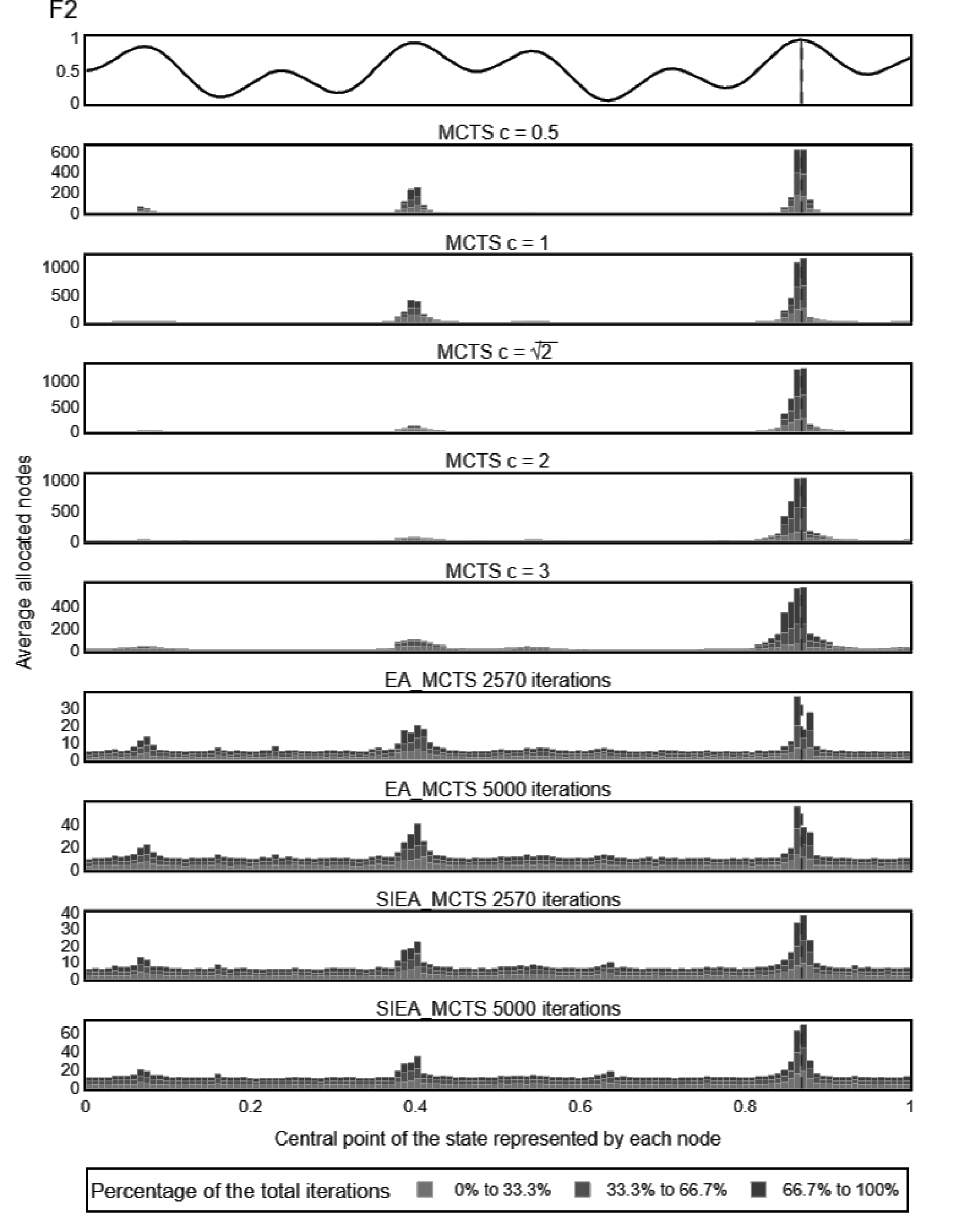}
   \caption{Histogram of the location of the nodes using MCTS UCT C = \{0.5,1,$\sqrt{2}$,2,3\} and SIEA-MCTS for f$_2$ (see Section~\ref{sec:background}).}

    \label{fig:f2}
\end{figure}

MCTS UCT behaves incredibly well when it is well-tuned such as using a `correct' $C$ constant value. The use of UCT $C=\sqrt2$ in MCTS has proven to perform well in multiple problems, although other $C$ values have also yielded good results~\cite{Fred}. Thus, in this study, {we use a range of UCT $C$ values used in MCTS and compare their search over 5000 MCTS iterations. This number of iterations is enough to allow MCTS to explore the search space as well as reaching the bottom of the tree given enough exploitation.}

{The EA-inspired MCTS variants (EA-MCTS and SIEA-MCTS) are given 5000 iterations from which 2430 are evolution-dedicated fitness iterations and 2570 are invested in tree search using the evolved formula. This setting models the usage of the EA-inspired MCTS variants evolving the UCT formula online. To fairly compare the behaviour of the evolved formulae to vanilla MCTS with UCT we make up for those fitness iterations in another set of experiments, where the EA-based methods are allowed a total of 7430 iterations (2430 fitness iterations plus 5000 tree search iterations). This is done to ensure that the EA-inspired MCTS variants have the same number of iterations to explore the search space as MCTS with UCT. }%The results of these experiments are shown in Figs.~\ref{fig:f1}--\ref{fig:f5}.}

{Three different stages of the search are tracked in each experiment to help us understand how the statistical tree grows over time. That is, 33\%, 66\% and 100\% of the total iterations. The first set of iterations is represented in light grey, the second set of iterations in dark grey and the final set of iterations is shown in black in Figs~\ref{fig:f1} -- \ref{fig:f5}, for f$_1$ to f$_5$, respectively (see Fig.~\ref{fig:functions}). On these figures, the $x$-axis pertains to the domain of the function, while the $y$-axis denotes the count of nodes expanded within a specific region (except the top subplot which displays the function being searched, as detailed in Section~\ref{sec:experimental}).} The results have been averaged over 30 independent runs for each AI controller employed in this study (see Section~\ref{sec:ai:controllers}). To construct the histograms, we organised the nodes of the tree into bins based on the location of the centre of their states. This approach aims to provide insights into how and when the nodes are expanded across different regions of the domain.

Let us start our analysis with the first function f$_1$ (refer to Section 3), conveniently re-plotted at the top of Fig.~\ref{fig:f1}. This function, characterised by a single peak, represents the easiest scenario, among the five functions used in this work, to find a solution. The five plots following the top one in Fig.~\ref{fig:f1}, from top to bottom, pertain to MCTS using different $C$ values, specifically $C=$\{0.5, 1, $\sqrt2$, 2, 3\}. It is evident that with an increase in the $C$ value, MCTS embarks on broader explorations. This contrast is particularly noticeable between the fifth plot, where $C = 3$, and the first plot, where $C = 0.5$. In the former, there are approximately 100 nodes around the global optimum, compared to nearly 300 nodes in the latter case.

Shifting our focus to our EA approach, we observe from the seventh plot onwards, from top to bottom, that the Semantically-Inspired Evolutionary Algorithm Monte Carlo Tree Search (SIEA-MCTS) methods exhibit a wider range of search variations compared to MCTS, irrespective of the UCT $C$ value applied. This outcome aligns with expectations, as in each node selection, we evolve an alternative UCT formula. Remarkably, our method demonstrates the ability to expand a significant number of nodes in close proximity to the global optimum (situated at the centre of f$_1$, as depicted at the top of Fig.~\ref{fig:f1}).  Upon closer examination of the outcomes generated by EA-MCTS, i.e., EAs without semantics, it becomes evident that this variant achieves a similar count of nodes surrounding the global optimum in comparison to the results yielded by SIEA-MCTS. The distinction in results between these two EAs methods lies in the fact that the EA-MCTS method samples a greater number of solutions in the `incorrect' part of the search space, as opposed to the outcomes produced by the SIEA-MCTS method.

Let us delve into our second function, as outlined in Section~\ref{sec:background}, displayed at the top of Fig.~\ref{fig:f2}. This function presents multiple peaks, with the global optimum situated on the right-hand side as represented by the vertical line. Analysing how MCTS performs with this function (observed in plots two to six from top to bottom in Fig.~\ref{fig:f2}), we note that as $C$ dramatically increases, MCTS demonstrates a tendency to expand nodes not only around the global optimum, but also within local optima. This behaviour is prominently depicted when $C=3$. As the UCT $C$ value decreases, this pattern shifts, except when $C = 0.5$, resulting in the formation of two peaks, with one coinciding with the location of the global optimum. Additionally, it is noteworthy that MCTS with UCT $C =$ \{1, $\sqrt2$\} exhibit similar behaviour: both expand nodes around the global optima region and explore a limited number of nodes within a local optima region.

Now, shifting our focus to the EA methods, we once again observe a broad spectrum of exploration and exploitation within the search space. Both EA methods effectively sample a relatively high number of nodes around the global optimum. Upon comparing both EA methods, we discern a trend: the SIEA-MCTS approach demonstrates a slightly greater propensity to sample points around the global optimum compared to EA-MCTS. We believe this distinction arises from the incorporation of semantics in SIEA-MCTS, which encourages diversity in the EA approach.

\begin{figure}[tb] 
    \centering\includegraphics[width=0.9\columnwidth]{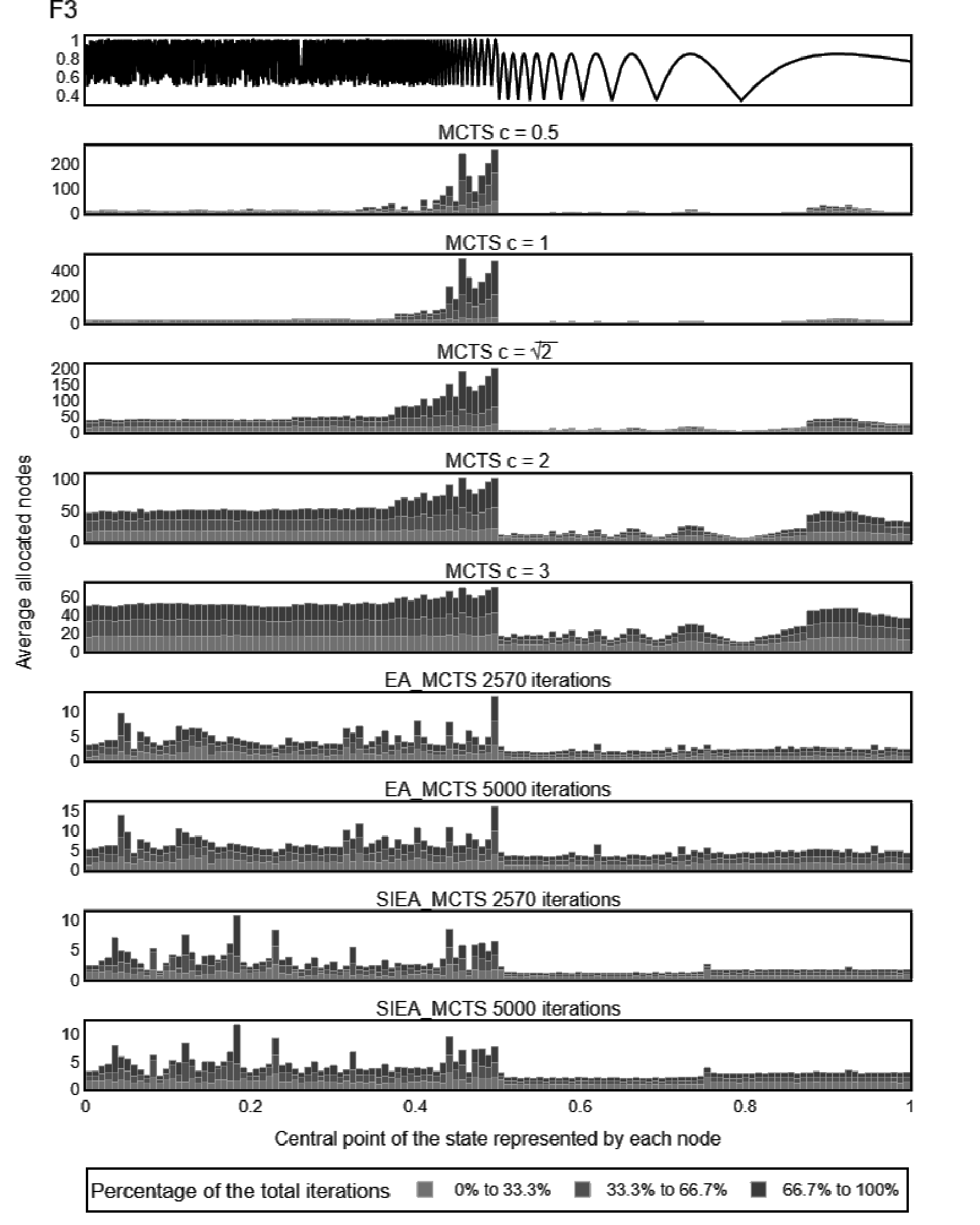}

    \caption{Histogram of the location of the nodes using MCTS UCT C = \{0.5,1,$\sqrt{2}$,2,3\} and SIEA-MCTS for f$_3$ (see Section~\ref{sec:background}).}
    \label{fig:f3}
\end{figure}

\begin{figure}[t] 
  \centering\includegraphics[width=0.9\columnwidth]{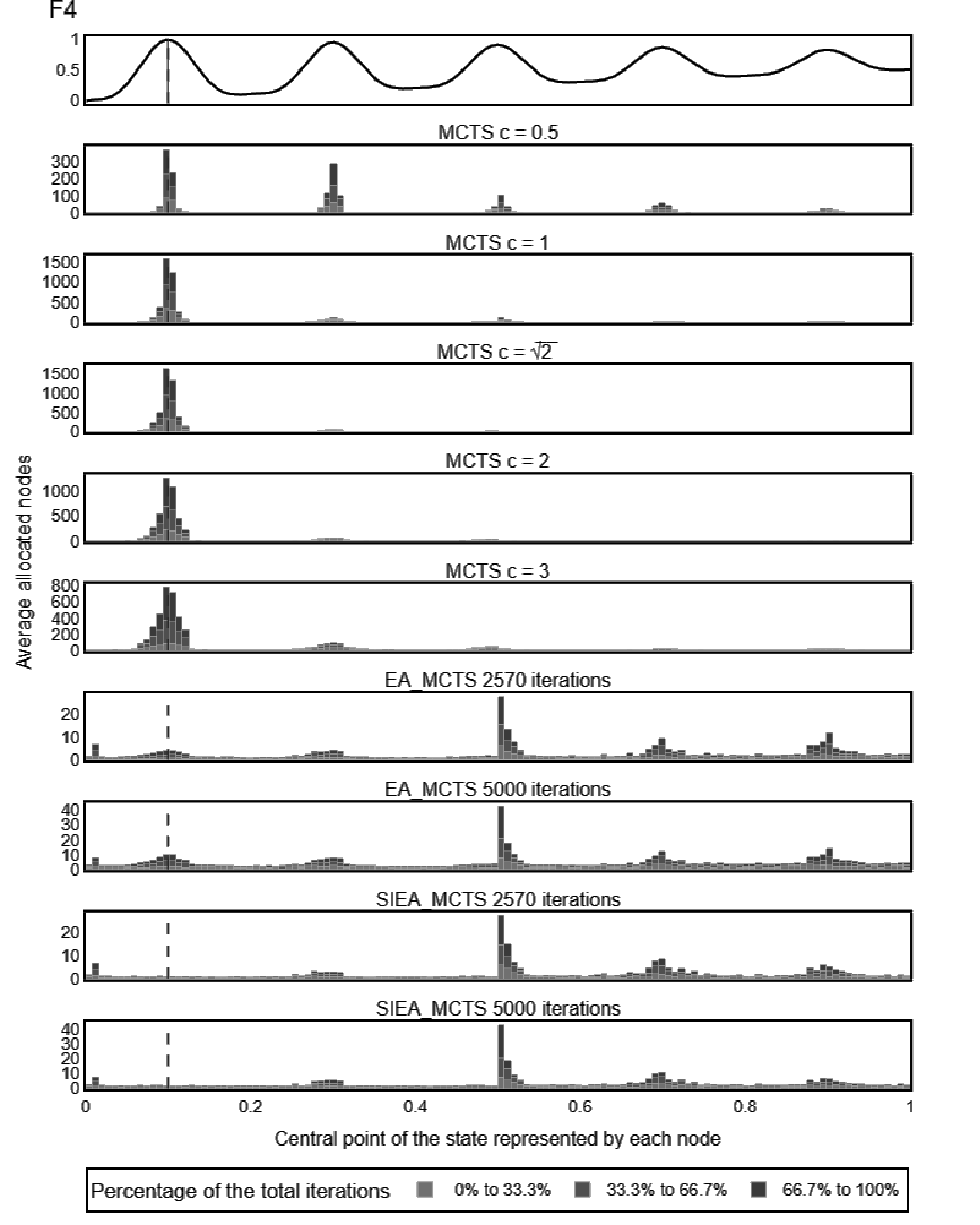}
   \caption{Histogram of the location of the nodes using MCTS UCT $C$= \{0.5,1,$\sqrt{2}$,2,3\}, and EAs and SIEA-MCTS for f$_4$ (see Section~\ref{sec:background}).}

    \label{fig:f4}
\end{figure}

\begin{figure}[t] 
    \centering\includegraphics[width=0.9\columnwidth]{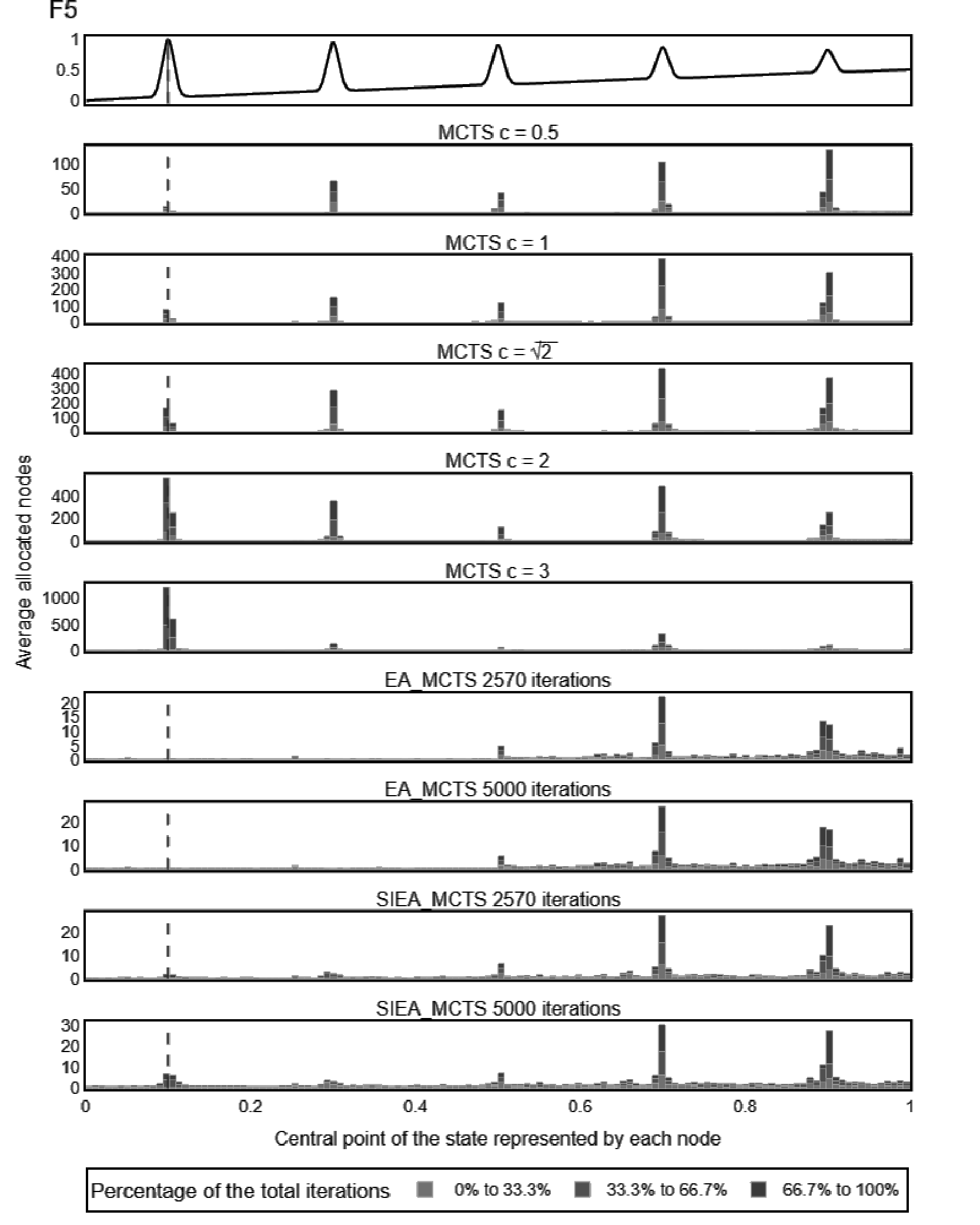}

     \caption{Histogram of the location of the nodes using MCTS UCT $C$ = \{0.5,1,$\sqrt{2}$,2,3\}, and  EAs and SIEA-MCTS for f$_5$ (see Section~\ref{sec:background}).}
    \label{fig:f5}
\end{figure} 

We now shift our focus to the third function used in this study (refer to Section~\ref{sec:background}), illustrated at the top of Figure~\ref{fig:f3}. This function exhibits a notable ruggedness, particularly evident on the left-hand side of the top plot, where multiple global optima are present. In stark contrast, the right-hand side of the function graph indicates a smoother behaviour. Examining the first five plots, barring the top plot, corresponding to MCTS UCT $C$ from top to bottom in Figure~\ref{fig:f3},  it becomes apparent that MCTS with a higher UCT $C$ value tends to explore more extensively. Consequently, the nodes are inclined to expand into `incorrect' regions of the search space (right-hand side of plots), as observed when $C = \{2, 3\}$. This phenomenon is also noticeable when $C = \sqrt2$. The scenario undergoes a significant shift as UCT $C$ decreases, as exemplified by the first plot where UCT $C = 0.5$. Here, a substantial number of nodes are expanded within a narrow region containing the global optimum.

Turning our attention to the behaviour of the EA (as depicted in the last four plots of Figure~\ref{fig:f3}), an interesting trend emerges: numerous peaks, albeit small, are evenly distributed in the correct region harbouring multiple global optima. This stands in stark contrast to the outcome observed when using MCTS with UCT $C = 0.5$, where some peaks manifest in only a fraction of the `correct' portion of the search space. When comparing the performance of EA-MCST and SIEA-MCTS, it is evident that the semantic-based EAs sample slightly fewer solutions in the incorrect part of the search space when compared to the EA-MCTS approach.

Let us turn our attention to the fourth function, as defined in Section~\ref{sec:background} and illustrated at the top of Figure~\ref{fig:f4}. This particular function is deceptive, with its global optimum situated on the left-hand side of the figure, denoted by a vertical line. Most MCTS variants perform commendably on this type of function. It is worth noting that MCTS with UCT $C = 1$ and $C = \sqrt{2}$ stand out, exhibiting the highest number of expanded nodes around the global optimum. Upon closer examination of the EA approach, intriguing behaviours emerge. Notably, EA-MCTS demonstrates the ability to sample points in proximity to the global optima, indicating an inherent capacity to evolve towards an optimal solution. However, it is fair to say that both EA method sample just a few solutions in the `correct' part of the search space. SIEA-MCTS displays similar behaviour to EA-MCTS in terms of point sampling across the search space.

Now, let us delve into the fifth and final function employed in this study, defined in Section~\ref{sec:background} and illustrated at the top of Figure~\ref{fig:f5}. This function poses a greater challenge compared to the previously examined fourth function, which was also deceptive, as well as the rest of the other functions. Unlike its predecessor f$_4$, this function f$_5$, demands more extensive exploration from MCTS; as $C$ increases, MCTS is more likely to uncover the global optima. Notably, the most effective MCTS variant is the one with $C = 3$ (as seen in the sixth plot from top to bottom). Shifting our focus to the performance exhibited by the EA methods, it becomes evident once again that the semantic-based approach excels in sampling points around the global optima. This stands in contrast to the behavior observed when using EA-MCTS. This reinforces the notion that semantics indeed fosters diversity, enabling SIEA-MCTS to outperform EA-MCTS.

From this analysis, we observe that MCTS UCT performs well in functions f$_1$ and f$_2$. It continues to demonstrate strong performance across the remaining functions, provided an appropriate UCT $C$ value is chosen, as discussed earlier. For instance, in the case of the deceptive function with some degree of smoothness (f$_4$), MCTS with UCT $C = 1$ and $C = \sqrt{2}$ exhibit commendable performance. However, for the more challenging deceptive function (f$_5$), MCTS with UCT $C = 3$ is the preferred choice.

Turning our attention to the EAs' behaviour in evolving an alternative formula, we find that the evolution of selection policies, as opposed to the use of UCT with tunned $C$ values, proves to be relatively effective across most functions. Nevertheless, it is worth noting that the EA method's sampling around the global optima is limited; only a few points are sampled around the global optima across all the functions studied in this work.

\subsection{Effects of the selection policy on the statistical tree}

To comprehensively grasp the implications of MCTS UCT in conjunction with the evolved selection policy across diverse search landscapes, as presented in Section~\ref{sec:background}, we define four metrics. The first two metrics, named (i) node expansion rate per iteration and (ii) number of terminal states reached, complement each other to describe the exploration/exploitation trade-off. The last two metrics, named (iii) most visited node-based result and (iv) highest reward-based result, inspired by~\cite{james2017analysis}, allow us to shed some light on the implications of selecting a node at the end of the iterations to get the best result possible, a long-term debate in the MCTS community.

{The first metric that we compute, node expansion rate per iteration, shown in the second column, from left-to-right in Tables~\ref{table:f1_results} --~\ref{table:f5_results} for f$_1$ -- f$_5$, respectively, is the percentage number of expanded nodes in the MCTS tree with respect to the total number of iterations executed by the algorithm. This metric will allow us to understand how explorative the MCTS is and  is principled  by the fact that an iteration will fail to expand a node on its expansion step if the selected node represents a terminal state, as no further nodes can be added to it. Our metric describes how often that was the case. A value of 1 (100\%) indicates that each MCTS iteration added a node to the tree on the expansion step, meaning that the exploitation was not enough to reach the bottom of the tree, hence demonstrating greater exploration.}

The second metric, number of terminal states reached, measures the number of unique terminal states added as nodes of the final statistical tree. The game tree of the FOP problem as defined in Section~\ref{sec:background} has a maximum depth of 17, which can  be reached by the tree search. This metric complements the first one by revealing how varied the terminal states are in the final statistical tree. If an agent has more than zero number of terminal stated reached it can be said that it explored less than any other agent equal to zero on this metric, as it implies that the search of the latter reached deeper in the tree than the first. The number of terminal states reached  is shown in the third column of Tables~\ref{table:f1_results} -- \ref{table:f5_results} for Functions 1--5, respectively

The third and fourth computed metrics, namely (iii) most visited node-based result  and (iv) highest reward-based result, serve to report the search results. The former is derived by traversing the tree path with the highest frequency of visits, while the latter is obtained by following the path with the highest reward values. These complementary metrics offer insight into which of these approaches yields superior results, thereby contributing to the ongoing discourse in the MCTS community regarding their relative merits. Furthermore, they provide a means to observe how these values may diverge based on the specific problem at hand, as we will discuss  next.

\begin{table}[tb]
\caption{Node expansion rate per iteration and number of terminal states reached (Cols. 2 and 3), and highest results based on most visited nodes and highest reward values (Cols. 4 and 5), for MCTS UCT using various $C$ values, and EAs and SIEA-MCTS (using 2570 and 5000 simulations -- read text), for f$_1$.}
\begin{tabular}{P{2.6cm}P{2.2cm}P{2.2cm}P{2.2cm}P{2.2cm}}
  %{@{}ccccc@{}}
  \toprule
   \textbf{MCTS variant} & \textbf{\PercentageNodesF} & \textbf{No. of terminal states reached} & \textbf{\MostVisitedF} & \textbf{\BestRewardF}\\   
   \midrule
   MCTS C=0.5&$1\pm0$&$0\pm0$&$1\pm0$&$1\pm0$\\
     MCTS C=1&$1\pm0$&$0\pm0$&$1\pm0$&$1\pm0$\\
            MCTS C=$\sqrt{2}$&$1\pm0$&$0\pm0$&$1\pm0$&$1\pm0$\\
           MCTS C=2&$1\pm0$&$0\pm0$&$1\pm0$&$1\pm0$\\
            MCTS C=3&$1\pm0$&$0\pm0$&$1\pm0$&$1\pm0$\\
            EA-MCTS (2570)&$0.52\pm0.47$&$74.55\pm154.47$&$0.89\pm0.18$&$0.76\pm0.31$\\
            EA-MCTS (5000)&$0.46\pm0.45$&$235.88\pm484.23$&$0.89\pm0.18$&$0.79\pm0.31$\\
            SIEA-MCTS (2570) &$0.49\pm0.47$&$60.05\pm139.45$&$0.89\pm0.18$&$0.84\pm0.26$\\
            SIEA-MCTS (5000) &$0.44\pm0.45$&$203.61\pm458.16$&$0.89\pm0.18$&$0.81\pm0.29$\\ 
\botrule
\end{tabular}
\label{table:f1_results}
\end{table}

\begin{table}[tb]
\caption{Node expansion rate per iteration and number of terminal states reached (Cols. 2 and 3), and highest results based on most visited nodes and highest reward values (Cols. 4 and 5), for MCTS UCT using various $C$ values, and EAs and SIEA-MCTS (using 2570 and 5000 simulations -- read text), for f$_2$.}
\begin{tabular}{P{2.6cm}P{2.2cm}P{2.2cm}P{2.2cm}P{2.2cm}}
  %{@{}ccccc@{}}
  \toprule
   \textbf{MCTS variant} & \textbf{\PercentageNodesF} & \textbf{No. of terminal states reached} & \textbf{\MostVisitedF} & \textbf{\BestRewardF}\\   
   \midrule
   MCTS C=0.5&$0.51\pm0.2$&$667.96\pm379.1$&$0.94\pm0.04$&$0.94\pm0.04$\\
                MCTS C=1&$0.96\pm0.04$&$1038.93\pm348.58$&$0.96\pm0.02$&$0.96\pm0.02$\\
                MCTS C=$\sqrt{2}$&$1\pm0$&$709.37\pm275.54$&$0.97\pm0.01$&$0.97\pm0.01$\\
                MCTS C=2&$1\pm0$&$180.85\pm136.42$&$0.97\pm0.01$&$0.97\pm0.01$\\
                MCTS C=3&$1\pm0$&$0\pm0$&$0.97\pm0.01$&$0.97\pm0.01$\\
                EA-MCTS (2570) &$0.33\pm0.43$&$19.94\pm47.63$&$0.75\pm0.25$&$0.69\pm0.25$\\
                EA-MCTS (5000) &$0.31\pm0.43$&$53.95\pm201.97$&$0.74\pm0.25$&$0.67\pm0.27$\\
                SIEA-MCTS 2570 &$0.38\pm0.45$&$16.81\pm48.18$&$0.77\pm0.22$&$0.67\pm0.29$\\
                SIEA-MCTS (5000)&$0.36\pm0.45$&$54.49\pm219.76$&$0.77\pm0.22$&$0.66\pm0.3$\\
                
\botrule
\end{tabular}
\label{table:f2_results}
\end{table}

 The values of the aforementioned four metrics for f$_1$, as elucidated in Section~\ref{sec:background} and illustrated in Fig.~\ref{fig:functions}, are presented in Table~\ref{table:f1_results}. For this unimodal function, the node expansion rate per iteration using MCTS UCT remains consistently at 1 (100\%), irrespective of the $C$ values defined, as indicated in the first column of Table~\ref{table:f1_results}. This aligns with previous findings~\cite{James_Konidaris_Rosman_2017}, which suggest that MCTS UCT tends to favour smooth areas, resulting in multiple leaf nodes exploring this region. Turning our attention to the number of terminal states reached  using MCTS UCT, we observe that none of them reach a terminal state (third column in Table~\ref{table:f1_results} for f$_1$), confirming the high exploratory nature of MCTS in this specific unimodal function.

 When focusing on the next set of functions, namely f$_2$, f$_3$, f$_4$, and f$_5$, representing smooth, rugged, and two deceptive landscapes, respectively (as depicted in Fig.~\ref{fig:functions}), and using the node expansion rate per iteration  metric, an interesting pattern emerges. This rate remains high when the MCTS UCT $C$ value is high, for instance, when $C=3$ (refer to Tables~\ref{table:f2_results} -- \ref{table:f4_results} for f$_2$, f$_3$, and f$_4$, respectively), {as is expected from the most exploratory versions of MCTS}. However, it is important to note that this value does not guarantee a 100\% expansion of nodes, as there is an exception for the highly deceptive function f$_5$, as illustrated in Table~\ref{table:f5_results}.

 Turning our attention to the number of terminal states reached for f$_2$, f$_3$, and f$_4$, we begin to observe distinctions in comparison to the zero count achieved by f$_1$ for expanded terminal nodes. It becomes evident that a high count is observed for f$_2$ when $C$ is low (see third column in  Table~\ref{table:f2_results}), and this stabilises as $C$ increases (specifically, for $C=3$). This same trend is observed for f$_3$ and f$_4$ (Tables~\ref{table:f3_results} and~\ref{table:f4_results}, respectively). In essence, the exploratory power of MCTS UCT starts to diminish as the problem complexity escalates, indicated by the count of expanded nodes for MCTS UCT in these functions (f$_2$, f$_3$, and f$_4$). This behaviour, however, is not observed in the highly deceptive f$_5$ employed in this study, where the number of expanded terminal nodes hovers around 700 nodes using the same $C$ = 3 (see Table~\ref{table:f5_results}).

Considering these two metrics, node expansion rate per iteration and number of terminal states reached, for the EA-based methods, denoted as SIEA-MCTS and EA-MCTS, we note that the node expansion rate per iteration hovers around 50\% (refer to the last four rows, second column in Tables~\ref{table:f1_results}--~\ref{table:f5_results}). This indicates that, in most cases, the evolved formulae obtained through these EAs methods tend to be `trivial', as they generally lack exploratory power, as evidenced by the low  node expansion rate per iteration. Furthermore, when examining the count of expanded terminal nodes, we find that the number displays considerable variation, which is to be expected given the diverse nature of evolved formulae.

\begin{table}[tb]
\caption{Node expansion rate per iteration and number of terminal states reached (Cols. 2 and 3), and highest results based on most visited nodes and highest reward values (Cols. 4 and 5), for MCTS UCT using various $C$ values, and EAs and SIEA-MCTS (using 2570 and 5000 simulations -- read text), for f$_3$.}
\begin{tabular}{P{2.6cm}P{2.2cm}P{2.2cm}P{2.2cm}P{2.2cm}}
  %{@{}ccccc@{}}
  \toprule
   \textbf{MCTS variant} & \textbf{\PercentageNodesF} & \textbf{No. of terminal states reached} & \textbf{\MostVisitedF} & \textbf{\BestRewardF}\\   
   \midrule
    MCTS C=0.5&$0.51\pm0.12$&$466.62\pm225.59$&$0.97\pm0.06$&$0.97\pm0.06$\\
            MCTS C=1&$1\pm0.01$&$132.05\pm124.66$&$0.99\pm0.01$&$0.99\pm0.02$\\
            MCTS C=$\sqrt{2}$&$1\pm0$&$0\pm0$&$0.98\pm0.05$&$0.98\pm0.05$\\
            MCTS C=2&$1\pm0$&$0\pm0$&$0.96\pm0.09$&$0.95\pm0.1$\\
            MCTS C=3&$1\pm0$&$0\pm0$&$0.91\pm0.13$&$0.91\pm0.13$\\
            EA-MCTS (2570) &$0.17\pm0.32$&$18.13\pm32.47$&$0.87\pm0.14$&$0.88\pm0.14$\\
            EA-MCTS (5000)&$0.15\pm0.32$&$22.12\pm61.98$&$0.88\pm0.13$&$0.87\pm0.14$\\
            SIEA-MCTS (2570) &$0.13\pm0.25$&$19.14\pm33.04$&$0.86\pm0.14$&$0.89\pm0.13$\\
            SIEA-MCTS (5000) &$0.09\pm0.24$&$19.26\pm33.41$&$0.87\pm0.14$&$0.89\pm0.12$\\
                
\botrule
\end{tabular}
\label{table:f3_results}
\end{table}

\begin{table}[tb]
\caption{Node expansion rate per iteration and number of terminal states reached (Cols. 2 and 3), and highest results based on most visited nodes and highest reward values (Cols. 4 and 5), for MCTS UCT using various $C$ values, and EAs and SIEA-MCTS (using 2570 and 5000 simulations -- read text), for f$_4$.}
\begin{tabular}{P{2.6cm}P{2.2cm}P{2.2cm}P{2.2cm}P{2.2cm}}
  %{@{}ccccc@{}}
  \toprule
   \textbf{MCTS variant} & \textbf{\PercentageNodesF} & \textbf{No. of terminal states reached} & \textbf{\MostVisitedF} & \textbf{\BestRewardF}\\   
   \midrule
   
          MCTS C=0.5&$0.35\pm0.19$&$430.33\pm354.38$&$0.91\pm0.06$&$0.91\pm0.06$\\
            MCTS C=1&$0.91\pm0.09$&$1168.34\pm310.68$&$0.97\pm0.02$&$0.97\pm0.02$\\
            MCTS C=$\sqrt{2}$&$1\pm0.01$&$991.72\pm246.33$&$0.98\pm0$&$0.98\pm0.01$\\
            MCTS C=2&$1\pm0$&$381.64\pm134.16$&$0.98\pm0$&$0.98\pm0$\\
            MCTS C=3&$1\pm0$&$0\pm0$&$0.98\pm0$&$0.98\pm0$\\
            EA-MCTS (2570) &$0.14\pm0.3$&$12.68\pm43.12$&$0.67\pm0.2$&$0.69\pm0.2$\\
            EA-MCTS (5000) &$0.12\pm0.28$&$22.28\pm86.01$&$0.67\pm0.2$&$0.68\pm0.21$\\
            SIEA-MCTS (2570)&$0.11\pm0.26$&$13.05\pm44.36$&$0.63\pm0.21$&$0.66\pm0.21$\\
            SIEA-MCTS (5000) &$0.09\pm0.24$&$23.47\pm88.42$&$0.63\pm0.21$&$0.66\pm0.21$\\       
\botrule
\end{tabular}
\label{table:f4_results}
\end{table}

\subsection{Node selection strategy: comparing the most visited path and the best reward path}

 {The most visited node-based result and highest reward-based result metrics, displayed in Columns 4 and 5 of Tables~\ref{table:f1_results} -- \ref{table:f5_results}, for f$_1$ to f$_5$, respectively, denote the agent's perceived value of the root state after optimal play~\cite{james2017analysis}. In MCTS, returning the most visited action as the recommendation policy is the most common approach taken by the community. However, given that the set of available actions conforms to a Multi-Armed Bandit model, the arm with the optimal reward is occasionally the more accurate choice~\cite{munos2014mabinmcts}. There is a contention that the most visited node-based result metric leans towards the exploitation of the most frequented path, which might not always be optimal. Conversely, the highest reward-based result  metric echoes recent discoveries of superior rewards, albeit with a higher uncertainty. Thus, we explore the results using these two metrics}.

As we defined the FOP in Section~\ref{sec:experimental}, the MCTS aims to maximise the reward. Consequently, the larger the most visited node-based result and highest reward-based result metrics are, the better MCTS's performance. {From Table~\ref{table:f1_results}, we can initially see that all vanilla MCTS agents (as represented in the first five rows) successfully achieved their search goals, incorporating the global optima into their statistical tree with a maximum reward. This was anticipated for the most basic and straightforward function. Conversely, the EA-based methods (as depicted in the last four rows) failed to locate the global optimum. Further, upon examining Tables~\ref{table:f2_results} -- \ref{table:f5_results}, it becomes apparent that vanilla MCTS variants consistently outperformed all EA-based MCTS variants across all functions. This performance discrepancy can be attributed to the inability of EA-based methods to sufficiently explore the search space to identify the global optimum. Additionally, the results presented are averaged over runs, where the EA struggled to evolve competitive formulae, as previously indicated by the low  no expansion rate per iteration  and number of terminal states reached values. However, there were instances where the least effective vanilla MCTS variant exhibited performance comparable to the EA-based methods (look, for example, to the highest reward-based result and most visited node-based result metrics for f$_3$ in Table~\ref{table:f3_results}). This implies that while the current version of the EA-based MCTS variants may not be robust enough, they can be competitive under specific conditions, especially when the domain necessitates a particular UCT $C$ value.
%% }

{Directing our attention to the first five rows of Tables~\ref{table:f2_results} -- \ref{table:f5_results}, which detail the performance of vanilla MCTS variants for functions f$_2$ through f$_5$, it is evident that distinct $C$ values yield superior outcomes for each function. This underscores the significance of finely calibrating the UCT $C$ value. Notably, the optimal results were achieved using the largest $C$ values for deceptive functions, as seen in  f$_4$ and f$_5$ (Tables~\ref{table:f4_results} and~\ref{table:f5_results}). This suggests that enhanced exploration is pivotal for identifying superior local or global optima in deceptive reward landscapes.
%% }

 {By comparing the most visited node-based result  and highest reward-based result  metrics from Tables~\ref{table:f1_results} -- \ref{table:f5_results}, a similarity emerges for the MCTS UCT variants across all functions. This outcome aligns with our expectations for our FOP, given that as we traverse further down the tree, the reward variance of the functions diminishes, and our iterations suffice to probe such depths. This consistency persists even for  f$_3$, notorious for its rugged reward landscape. In stark contrast, the EA-based methods display disparities between the most visited node-based result  and highest reward-based result metrics. For every EA-based MCTS variant, the most visited node-based result  value (third column) surpasses the highest reward-based result value (fourth column) for functions f$_1$ and f$_2$, which are unimodal and smooth, respectively. Conversely, for deceptive functions f$_4$ and f$_5$, the most visited node-based result  value lags behind the highest reward-based result. These observations insinuate that the optimal recommendation policy—whether favouring the most visited or the best reward—hinges on both the intrinsic attributes of the problem and the traits of the tree search agent.
%% }

 Focusing on the EA-based methods, it is noteworthy to mention the pronounced variance in their most visited node-based result  and highest reward-based result metrics, indicating a mix of successful and unsuccessful runs. Furthermore, there is not a discernible augmentation in either most visited node-based result  or highest reward-based result, even with an escalation in iteration numbers for any of the EA-based methods. For illustration, consider the last two rows of Table~\ref{table:f4_results} (f$_4$, deceptive function), where the Semantically-Inspired Evolutionary Algorithm Monte Carlo Tree Search (SIEA-MCTS)  agent's performance remained unchanged with 2570 or 5000 tree search-dedicated iterations. The underlying reason is that a significant proportion of the evolved formulae remain entrenched, capitalising on identical search space regions, rendering additional iterations redundant. This is corroborated by their subdued  node expansion rate per iteration and number of terminal states reached values, as discussed in the previous section. Consequently, one might deduce that while EA-based methods exhibit potential, they currently lack robustness.

\begin{table}[tb]
\caption{Node expansion rate per iteration and number of terminal states reached (Cols. 2 and 3), and highest results based on most visited nodes and highest reward values (Cols. 4 and 5), for MCTS UCT using various $C$ values, and EAs and SIEA-MCTS (using 2570 and 5000 simulations -- read text), for f$_5$.}
\begin{tabular}{P{2.6cm}P{2.2cm}P{2.2cm}P{2.2cm}P{2.2cm}}
  %{@{}ccccc@{}}
  \toprule
   \textbf{MCTS variant} & \textbf{\PercentageNodesF} & \textbf{No. of terminal states reached} & \textbf{\MostVisitedF} & \textbf{\BestRewardF}\\   
   \midrule
    MCTS C=0.5&$0.11\pm0.03$&$113.68\pm80.22$&$0.82\pm0.08$&$0.82\pm0.08$\\
            MCTS C=1&$0.3\pm0.06$&$359.65\pm131.91$&$0.86\pm0.05$&$0.86\pm0.05$\\
            MCTS C=$\sqrt{2}$&$0.45\pm0.06$&$533.96\pm149.6$&$0.87\pm0.06$&$0.87\pm0.06$\\
            MCTS C=2&$0.65\pm0.07$&$772.21\pm128.72$&$0.91\pm0.06$&$0.91\pm0.06$\\
            MCTS C=3&$0.9\pm0.07$&$698.98\pm208.6$&$0.95\pm0.05$&$0.95\pm0.06$\\
            EA-MCTS (2570) &$0.07\pm0.18$&$12.24\pm37.26$&$0.49\pm0.22$&$0.51\pm0.19$\\
            EA-MCTS (5000) &$0.05\pm0.17$&$15.09\pm49.6$&$0.49\pm0.23$&$0.51\pm0.19$\\
            SIEA-MCTS (2570) &$0.09\pm0.22$&$15.19\pm48.02$&$0.5\pm0.26$&$0.54\pm0.2$\\
            SIEA-MCTS (5000) &$0.07\pm0.2$&$18.16\pm63.37$&$0.49\pm0.27$&$0.54\pm0.21$\\
                
\botrule
\end{tabular}
\label{table:f5_results}
\end{table}

\section{Conclusions}
\label{sec:conclusions}

MCTS is a best-first sampling method employed in the search for optimal decisions. It has yielded extraordinary results in challenging problems such as the Game of Go, a long-standing problem in AI, where methods relying on MCTS have beaten master Go players. The effectiveness of MCTS relies on the construction of its statistical tree, with the selection policy playing a fundamental role. Multiple selection policies have been proposed by the research community to be used in MCTS, with the Upper Confidence Bounds for Trees, referred to as UCT, being the most popular thanks to its incredible performance.

It is, however, worth noting that the MCTS UCT requires some tuning to perform well on different tasks. As a result of this, multiple scientific works have proposed automatically evolving the UCT through EAs. Most of these works have focused their attention on a single problem, impeding the generalisation of their findings. In sharp contrast, this work has used five different functions of varying difficulties and features, allowing us to shed light on the opportunities and limitations of MCTS UCT as well as its evolved selection policies. We have learned that MCTS UCT performs well in all the functions used in this study, including unimodal, multimodal, and deceptive functions. However, in all of these cases, it was necessary to tune the UCT to achieve a well-performing MCTS search. This limitation can be addressed by evolving the MCTS UCT through the use of EAs. We have seen how a semantically inspired EA is able to produce well-behaved evolved formulae to be used instead of UCT in MCTS. These formulae, used in lieu of the UCT in MCTS, are able to sample solutions in the correct part of the search space, regardless of the type of landscape represented by these five functions, as discussed in Section~\ref{sec:results}. It is, however, fair to say that with enough simulations and correct tuning of the MCTS UCT, it performs better compared to those solutions evolved through EAs.

\section*{Acknowledgments}

        This work has emanated from research conducted with the financial support of Science Foundation Ireland (SFI) under Grant Number SFI 18/CRT/6049. The authors wish to acknowledge the Irish Centre for High-End Computing (ICHEC) for the provision of computational facilities and support. This research paper has been meticulously developed through collaborative efforts among the authors, involving extensive research including implementing our own code and running experiments, fruitful discussions, and more. To ensure a high standard of presentation, professional software tools have been employed to craft visually appealing figures (e.g., CorelDraw) and enhance the overall polish of the document (e.g., ispell, ChatGPT) instead of automatically generating text.

%\bibliography{sn-bibliography}% common bib file
\bibliography{fop_mcts_eas_v2}
%% if required, the content of .bbl file can be included here once bbl is generated
%%\input sn-article.bbl

\end{document}